\documentclass[conference]{IEEEtran}

\IEEEoverridecommandlockouts

\usepackage{cite}
\usepackage{amsmath,amssymb,amsfonts}
\usepackage{algorithm}
\usepackage{algpseudocode}
\usepackage{textcomp}
\usepackage{xcolor}

\providecommand{\doi}[1]{doi: {\footnotesize \href{http://dx.doi.org/#1}{\path{#1}}}}

\usepackage[pdftex=true,breaklinks=true,hidelinks=true,colorlinks=false,citecolor=blue]{hyperref}

\usepackage{graphicx}
\graphicspath{{pdf/}{photo/}{figures/}}
\DeclareGraphicsExtensions{.eps,.jpeg,.jpg,.png,.pdf}

\usepackage{subcaption}
\usepackage[belowskip=-16pt,aboveskip=1pt]{caption}

\usepackage{fancyhdr}
\usepackage{booktabs}
\usepackage{threeparttable}

\usepackage[a4paper, hmargin=13mm, vmargin={25mm,28mm}, headsep=5mm]{geometry}

\usepackage[font=footnotesize]{caption}

\def\BibTeX{{\rm B\kern-.05em{\sc i\kern-.025em b}\kern-.08em
    T\kern-.1667em\lower.7ex\hbox{E}\kern-.125emX}}
\begin{document}

\makeatletter
\fancypagestyle{plain}{
    \fancyhf{}
    \fancyhead[C]{\emph{Preprint submitted to the 14\textsuperscript{th} Mediterranean Conference on Embedded Computing,  10-14 June 2025, Budva, Montenegro}}
    \fancyfoot[C]{\thepage}
     \renewcommand{\headrulewidth}{0pt}
}
\makeatother

\makeatletter
\fancypagestyle{custom}{
    \fancyhf{}
    \fancyfoot[C]{\thepage}
    \renewcommand{\headrulewidth}{0pt}
}
\makeatother

\title{Transforming Faces Into Video Stories-VideoFace2.0\\
\thanks{\IEEEauthorrefmark{4} equal contribution}
}

\author{
\IEEEauthorblockN{Branko Brkljač\textsuperscript{1}\IEEEauthorrefmark{4}\IEEEauthorrefmark{1}, Vladimir Kalušev\textsuperscript{2}\IEEEauthorrefmark{4}, Branislav Popović\textsuperscript{1}, Milan Sečujski\textsuperscript{1}}
\IEEEauthorblockA{
\textsuperscript{1}\textit{\fontsize{14}{12}Department of Power, Electronic and Telecommunication Engineering},
\textit{Faculty of Technical Sciences, University of Novi Sad}\\
Trg Dositeja Obradovića 6, 21000 Novi Sad, Serbia\\
{\fontsize{9}{9}{\{}{{brkljacb}\IEEEauthorrefmark{1}}{, bpopovic, secujski}{\}}{@uns.ac.rs}}
}
\IEEEauthorblockA{
\textsuperscript{2}\textit{\fontsize{14}{12}Visual Computing \& Perception Group},
\textit{The Institute for Artificial Intelligence Research and Development of Serbia}\\
Fruškogorska 1, 21000 Novi Sad, Serbia\\
{\fontsize{9}{9}{vladimir.kalusev@ivi.ac.rs}}
}
}

\maketitle
\thispagestyle{plain} 

\begin{abstract}
Face detection and face recognition have been in the focus of vision community since the very beginnings. Inspired by the success of the original Videoface digitizer, a pioneering device that allowed users to capture video signals from any source, we have designed an advanced video analytics tool to efficiently create structured video stories, i.e. identity-based information catalogs. VideoFace2.0 is the name of the developed system for spatial and temporal localization of each unique face in the input video, i.e. face re-identification (ReID), which also allows their cataloging, characterization and creation of structured video outputs for later downstream tasks. Developed near real-time solution is primarily designed to be utilized in application scenarios involving TV production, media analysis, and as an efficient tool for creating large video datasets necessary for training machine learning (ML) models in challenging vision tasks such as lip reading and multimodal speech recognition. Conducted experiments confirm applicability of the proposed face ReID algorithm that is combining the concepts of face detection, face recognition and passive tracking-by-detection in order to achieve robust and efficient face ReID. The system is envisioned as a compact and modular extensions of the existing video production equipment. Presented results are based on test implementation that achieves between 18--25 fps on consumer type notebook. Ablation experiments also confirmed that the proposed algorithm brings relative gain in the reduction of number of false identities in the range of 73\%--93\%.
\end{abstract}


\section{Introduction}
\label{Introduction}
\pagestyle{custom}

\begin{figure}[!t]
\centering
    \includegraphics[width=0.98\columnwidth]{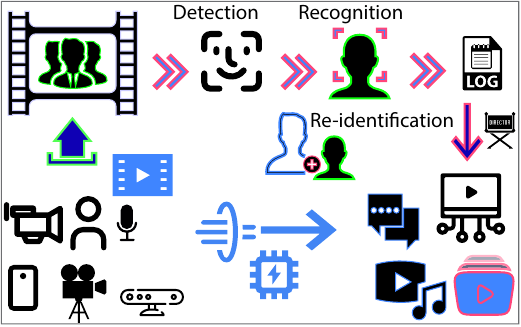}
   \captionsetup{width=0.98\linewidth}
\caption{VideoFace~2.0 processing workflow and applications.}
\label{fig:VideoFace2.0}
\end{figure}

Video content has become the prevailing type of global internet traffic \cite{ciscoVNI2019}. Similarly, media productions that include news, podcasts, and interviews had been steadily growing in the new internet formats. Due to the large amount of available data, different types of advanced machine learning (ML) vision tasks involving human speakers had also come into the focus of research community. However, availability of adequate video analysis tools represents a technological burden for more effective mastering of the respective tasks. Inspired by the historical success of the original Videoface digitizer \cite{videoface1987} we have developed  VideoFace2.0, Fig.~\ref{fig:VideoFace2.0}. The main assumption is that analyzed video content, regardless of source and application, contains appearance of multiple persons or speakers that randomly enter the scene  during the video, with face orientation towards the camera. The main goal of the system is to efficiently perform spatial and temporal localization of each unique face in the input video, i.e. solve the face re-identification (ReID) task and provide functionalities for cataloging, characterization and production of structured video outputs or video stories, Fig.~\ref{fig:VideoFace2.0}. It is assumed that neither the identities nor the number of people are known in advance.

Input video from recorded file, network stream or camera feed is uploaded into the system and analyzed frame-by frame in order to detect human faces (spatial and temporal localization), catalog their identity (create a new entry, i.e. video story or re-identify the person if it was already present in the video before). It can also generate additional personal characteristics with specific time stamps (e.g. facial landmarks, facial dynamics, age, gender and audio/speech). The system is designed to operate in near real-time mode and allows for standalone embedded hardware implementation.

VideoFace2.0 is primarily designed to be utilized in application scenarios that include TV production, media analysis and creative industries, where video stories generated by the system would be an efficient way for fast and easier video montage of scenes in which specific persons of interest appear. E.g. editing of interviews or reportages brought by the field reporters to TV studios, talk-shows, podcasts and other formats that include multiple speakers or participants.

Another important application field includes production of custom video-based datasets that are required  in different ML tasks involving multi-modal inputs like speech, text and image.  Challenging vision tasks like lip reading and multi-modal speech recognition require large-scale language and speaker specific video datasets necessary for ML model training. Thus, one of the main goals of VideoFace2.0 and the proposed functionality and designed elements was also to facilitate creation of high-quality multi-modal datasets involving human speakers. The main contribution of the paper is the  generic and modular algorithm for face ReID, which is combining the concepts of face detection, recognition and passive tracking-by-detection.

The rest of the paper is organized as follows.
In Section~\ref{Related work} we introduce the main vision tasks behind the proposed processing workflow and discuss current state-of-the-art in the field. Next, in Section~\ref{Method and implementation} we present the main technical and algorithmic elements of the proposed solution, including proposed person re-identification improvements. Section~\ref{Results and discussion} brings results of conducted experiments, performance evaluation and discussion of possible limitations. Finally, in Section~\ref{Conclusions} we reexamine the strategies for further improvements and provide a reference for future work and planned applications. Code implementation is available at: {\url{https://github.com/brkljac/VideoFace2.0}}, while the results of test experiments are also available on the following link: \href{https://youtube.com/playlist?list=PLrYDQtLBCliyMwlyNJ62EtHiQYDiPxZlH&si=jtcxBqNCL3-Q8ZDW}{https://youtube.com/@kalusev}.

\section{Related work}
\label{Related work}
The proposed processing workflow, Fig.~\ref{fig:VideoFace2.0}, is based on four vision tasks that have been extensively studied and analyzed in the literature. These are face detection \cite{viola2004robust, zafeiriou2015survey}, face recognition \cite{wright2008robust, wang2023survey}, face re-identification (ReID) \cite{cheng2020face} and object tracking. Since each of these tasks is characterized by specific performance metrics and comprehensive evaluation benchmarks that go beyond the scope of this presentation, in this paper we have decided to rely on off-the-shelf solutions that have been properly tested and  proven as competitive to current state-of-the-art approaches in practice. In terms of applied algorithms and chosen pre-trained models we have followed the usual paradigm in which face detection results are used as a prerequisite for face characterization using learned face embeddings and followed by face identity matching in the corresponding high dimensional feature space during the face ReID phase. Nevertheless, we provide a brief overview of related works that illustrate design challenges and characteristics of the proposed VideoFace2.0.

The main goal of face detection is to provide localization (bounding box) and probability (decision confidence score)  of face appearance in the image. Since design requirements almost always are that face detection algorithm are fast and lightweight in terms of computational and memory resources, detection is usually performed over still images or individual video frames without the use of temporal information. It is a special case of object detection task, and as such depending on application requires the choice of optimal working point on the detector's receiver operating characteristic \cite{whalen1971detection}. The trade-off between face detection sensitivity and the level of false alarms (precision) is determined by detector's role in processing workflow. Similarly to other object detectors \cite{dollar2011pedestrian}, traditionally face detection was also designed as a two stage process involving face localization (e.g. sliding window search or region proposal networks \cite{ren2016faster}) and subsequent binary classification of region of interest (ROI). However, this changed with introduction of more efficient single stage detectors \cite{redmon2016you, deng2020retinaface}.

Open-set face ReID task is a challenging vision problem that is attracting significant attention of research community. It can be regarded as an extension of face recognition task, in which face identity is determined by matching the query image of the face with collected face samples in the dynamically changing gallery or database of face identities. ReID or recognition can be based on metric learning, i.e. query association with the closest identity in the gallery based on nonlinear distance between their vector embeddings. Embedding computation is usually preceded by face alignment \cite{onaran2024impact}, which is also the case in this paper. Since pre-trained feature extraction models can be obtained using different learning strategies, it creates possibility to impose various additional requirements on embedding invariance to different signal perturbations or distortions.

A closely related problem to face ReID is also object tracking in video. In case of active tracker, after initialization generally there is no need for subsequent object detection or recognition in following frames, but the matching of ROIs can still be based on described nonlinear image embeddings. However, in specific setups that are characterized by efficient face detection models, active tracking can be easily replaced by equally successful passive tracker \cite{bochinski2017high}, which is significantly less computationally expensive. This is exactly the design approach that was pursued in the proposed VideoFace2.0, where the tracker component is introduced in order to alleviate possible identity mismatch. E.g. when the person that was present in the previous frame is not correctly recognized (paired with  generated face embedding of any available identity in the gallery during the recognition phase), or not present (tracker confirms the new identity).

\section{Method and implementation}
\label{Method and implementation}
During the design process we have relied on gradual system improvement and synergistic effect of combining different vision components described in Section~\ref{Related work}. The primary requirements were robust face based identification of all persons in the video and at the same time efficient implementation with near real-time processing. The main elements of the proposed face re-identification method are summarized in  Algorithm~\ref{alg:faceReID}.

\begin{algorithm}
\scriptsize
\caption{}
\begin{algorithmic}[1]
    \Procedure{\textsc{VideoFace2.0 face ReID}}{}
        \State \textbf{\textit{Initialization:}} \textit{$I(t)$,  video frame  at time $t$; open-set face gallery ${\cal{G}}$; face detector~${\cal{D}}$ and confidence threshold $\sigma_h$; face embedding ${\cal{R}}$; distance $d(\cdot,)$ and threshold $\tau_d$; passive tracker ${\cal{T}}$ and IoU threshold $\tau$; post filtering period $t_{min}$;  }
       \State  Detect new faces: \textit{${\cal{F}}={\cal{D}}(I(t)) = \{F_1,  ..., F_N\}$}
         \State \textbf{\textit{Determine identities:}}
         \State  Discard unreliable $F_i$: ${\cal{F}}={\cal{F}}\setminus\{ \sigma(F_i)<\sigma_h \}, i\in1..N$
         \State  Recognition:
        \State \textbf{\textit{\# 1:}}\;\textit{compute ${\cal{R}}_{\cal{F}} = {\cal{R}}({\cal{F}}) = \{{\cal{R}}_{F_1},  ..., {\cal{R}}_{F_N}\}$ }

         \State \textbf{\textit{\# 2:}}\textit{$\;\forall \,{\cal{R}}_{F_i} \in {\cal{R}}_{\cal{F}}$ $\wedge$ ${\cal{R}}_{\cal{G}}=  \{{\cal{R}}_{G_1},  ..., {\cal{R}}_{G_K}\}$ \textbf{\textit{do:}}}

          \State \textbf{\textit{\# 3:}}\textit{ \textbf{\textit{if:}}  $\min_{j=1..K} d({\cal{R}}_{F_i}, {\cal{R}}_{G_j} ) < \tau_d$ \\ $\quad\quad\quad\;$ \textbf{\textit{then:}} new detection $i$ is re-identified as an existing person $j^{*} = \arg\min_{j=1..K}  d({\cal{R}}_{F_i}, {\cal{R}}_{G_j} )$ }

          \State \textbf{\textit{\# 4:}}\textit{\;\textbf{\textit{else:}} new detection "$i$" is not significantly similar to any of $K$ existing identities in gallery ${\cal{G}} \Rightarrow$ possibly new identity (candidate for inclusion in gallery ${\cal{G}}$): \textbf{call} ${\cal{T}}$}

          \State \textbf{\textit{\# 4.1:}}\textit{\;for candidate "$i$" call tracker ${\cal{T}}$ and check $IoU$ between bounding box of the closest detection $G_c \in {\cal{G}}$ and bounding box of $F_i$. \textbf{\textit{if:}} ${\cal{T}}_{IoU}(F_i, G_c)<\tau$ \textbf{\textit{then:}} new candidate identity is not valid, \textbf{\textit{else:}} add $F_i$ as new identity to ${\cal{G}}$:  ${\cal{G}}={\cal{G}}\cup{{F_i}}$ $\wedge$ $ {\cal{R}}_{\cal{G}}= {\cal{R}}_{\cal{G}}\cup{\cal{R}}({{F_i}})$, $|{\cal{G}}|=K+1$ }

          \State \textbf{\textit{Post filtering:}}\textit{$\;$If $t_{min}>0$ , next $t_{min}$ frames count number of appearances of all newly added identities at step $t$, and only after that make them active in the gallery (i.e. put them on hold for $t_{min}$ to confirm their validity). Otherwise discard them from the gallery.}
    \EndProcedure
\end{algorithmic}
    \label{alg:faceReID}
\end{algorithm}


Starting assumption of the adopted fast prototyping approach was that each of the face detection, recognition and tracking algorithms in Section~\ref{Related work} could be very efficient and stand out individually,  but that on their own they would not provide the desired level of face identification functionality. This was confirmed through preliminary tests, where despite the high level of sophistication and performance each of the above-mentioned components individually failed to provide consistent and robust extraction of faces from the video. E.g. face detector performance used to vary depending on sensitivity to input video signal characteristics. Face recognition that relied on pre-trained face embedding models was exhibiting high level of mismatches due to the fact that the embeddings were learned on unoccluded or less diverse face images compared to those present in the open-set of real-world human face appearances. Similarly, tracker component had difficulty in maintaining precise position and extent of the face over the entire video, with unresolved interruptions in the event that a person goes out of frame. On the other hand, design approaches that have proven the concept of combining highly efficient individual components in order to achieve functionality that goes beyond their individual performance, like the passive tracker in \cite{bochinski2017high}, were the main motivation for the proposed face ReID method.

The main concept of Algorithm~\ref{alg:faceReID} is that in terms of performance and efficient implementation face detection is the most mature vision technology in Fig.~\ref{fig:VideoFace2.0}, and as such should be the core of the proposed system. Thus, the underlying assumption was that the chosen detector ${\cal{D}}$ will work in high sensitivity mode with possibly high false positives rate per frame $I(t)$, but very unlikely miss of the face $F_i$ at time $t$ if detection $i$ has confidence score $\sigma(F_i)>\sigma_h$. On the other hand, matching of face inside the bounding box of $F_i$ with one of the already present open-set identities in gallery ${\cal{G}}=\{G_1, ..., G_K\}$ is based on computation of similarity or distance score $d(\cdot, \cdot)$ defined by eq.~\eqref{eq:cosine_distance}. It operates on high-dimensional feature vectors $\mathbb{R}^{d}$ generated by nonlinear embedding ${\cal{R}}$  that is pre-trained on face recognition task.

\begin{equation}
\label{eq:cosine_distance}
\begin{aligned}
\cos\angle({\cal{R}}_{F_i}, {\cal{R}}_{G_j}) &= \frac{{\cal{R}}_{F_i} \cdot {\cal{R}}_{G_j}}{\|{\cal{R}}_{F_i}\| \|{\cal{R}}_{G_j}\|}, \\
d({\cal{R}}_{F_i}, {\cal{R}}_{G_j}) &= 1 - \cos\angle({\cal{R}}_{F_i}, {\cal{R}}_{G_j}).
\end{aligned}
\end{equation}

Thus, if the matching of ${\cal{R}}(F_i)={\cal{R}}_{F_i}$ with any of $K$ known feature vectors ${\cal{R}}_{G_j}\in\mathbb{R}^{d}, j=1..K$ in gallery fails, it is an indicator of the possibly new face identity. However, since it is assumed that the adopted embedding ${\cal{R}}$ is pre-trained on some large-scale but standard face recognition task, despite being highly discriminative it will be exhibiting poor identity matching performance, i.e. high identity rejection rate, under considered open world setting. Namely, the presence of varying facial expressions, head pose, illumination, face occlusions (by person's hands or other objects like microphone or glasses) will lead to situations in which embeddings ${\cal{R}}_{F_i}, {\cal{R}}_{G_j}\in\mathbb{R}^{d}$, despite being very informative (with high discriminative power), will lead to undesirable rejections of face detections $F_i$ belonging to already known identity $j^{*}$ in the gallery. It will happen due to relatively simple decision rule, which is based on combining with cost efficient distance functions like cosine similarity distance $d$ in~\eqref{eq:cosine_distance}. Therefore, in order to avoid possible false rejections of known identities, ReID threshold $\tau_d$ for distance based test statistic in Algorithm~\ref{alg:faceReID} was intentionally raised.

In order to further improve ReID performance, additional passive tracker ${\cal{T}}$ was also included in Algorithm~\ref{alg:faceReID}, however in a careful way that does not increase processing latency. It means that it was designed not to be constantly active,  but only when needed, in addition to being computationally light (tracking-by-detection). Namely, since the detector and recognizer are successfully complementing each other most of the time, the only situations in which the tracker is invoked are when additional confirmation of the detection of new identity is necessary. In such cases, standard Intersection-over-Union (IoU) metric and the chosen IoU threshold $\tau$ are utilized to confirm that new identity has entered the video, step 4.1 in Algorithm~\ref{alg:faceReID}. We note that such setting was possible since in the planned application scenarios, Fig.~\ref{fig:VideoFace2.0}, faces are expected to enter the scene gradually, and it is highly unlikely that a new face will appear immediately over the same position of some face from the previous frame.

In addition to the described algorithmic elements, there is also a final, post filtering step where the  deliberately introduced delay $t_{min}$ can be traded for additional protection against fake new identities. However, introduced $t_{min}$ delay in new identity approval only affects new identities when they appear for the first time, and could be replaced by a more complex ReID rule.

\section{Results and discussion}
\label{Results and discussion}

\begin{figure}[tb]
  \centering
  \begin{subfigure}{0.3\columnwidth}
    \centering
    \includegraphics[width=\linewidth]{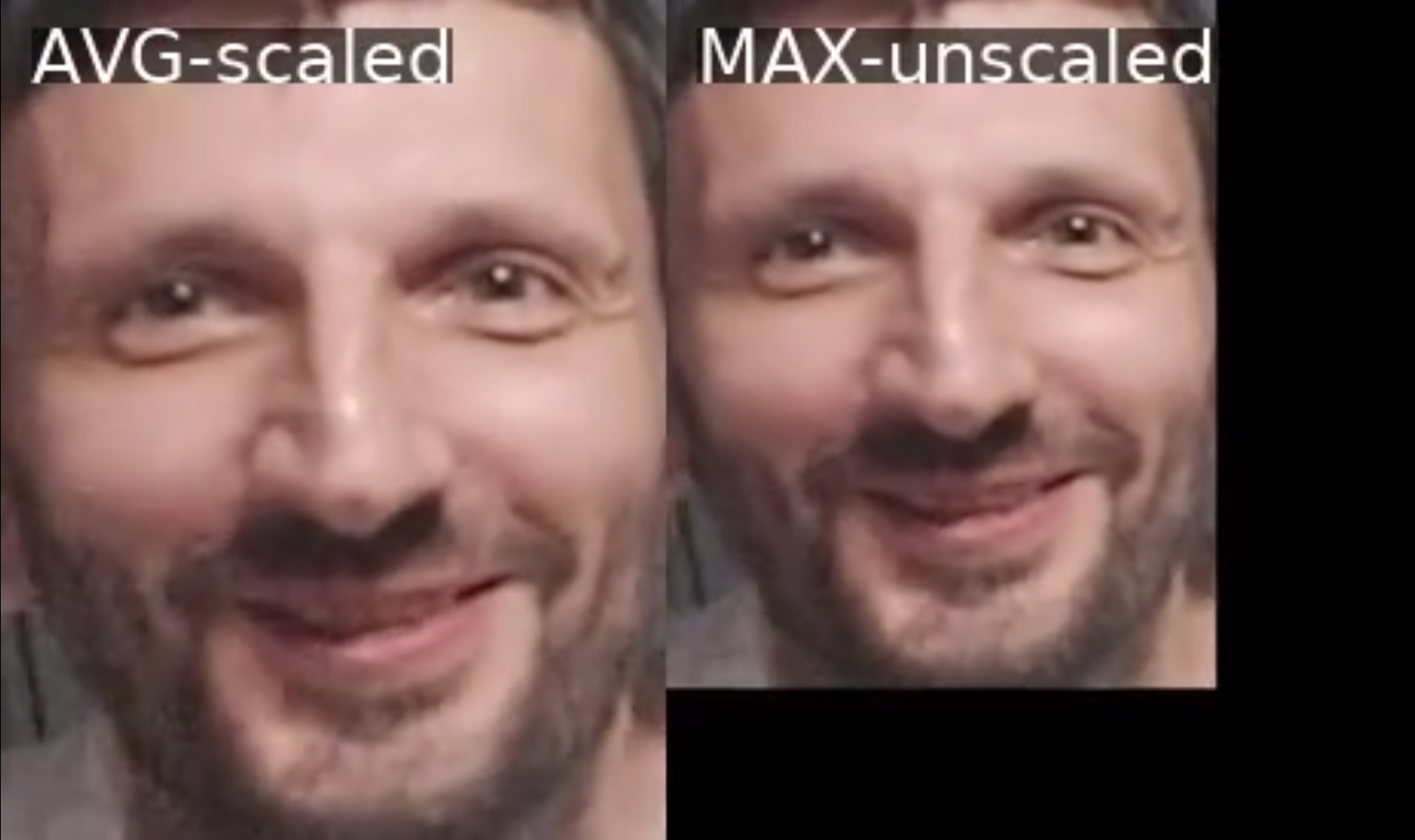}
    \caption{}
    \label{fig:fig2a}
  \end{subfigure}\hfill
  \begin{subfigure}{0.25\columnwidth}
    \centering
    \includegraphics[width=\linewidth]{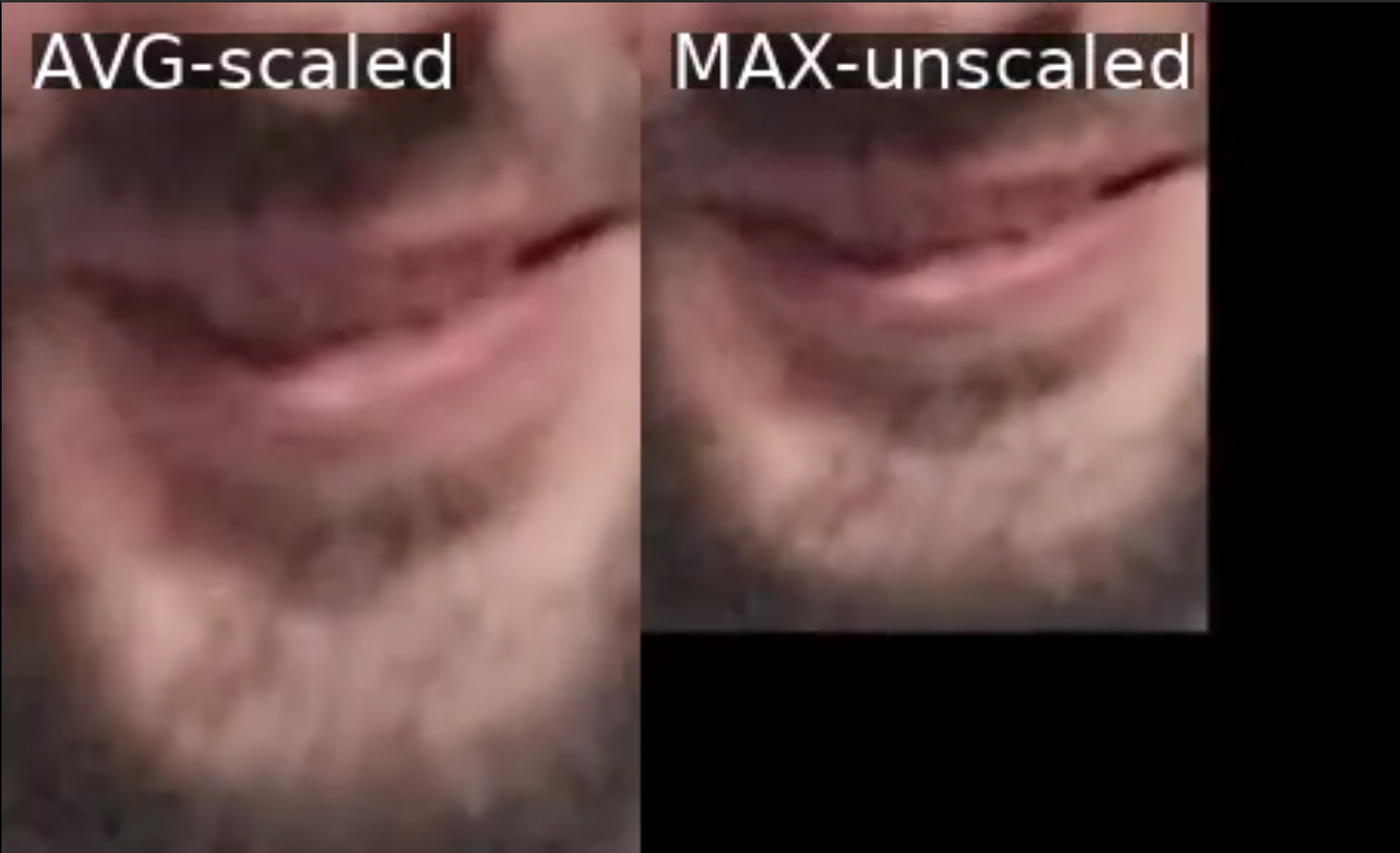}
    \caption{}
    \label{fig:fig2b}
  \end{subfigure}\hfill
  \begin{subfigure}{0.3\columnwidth}
    \centering
    \includegraphics[width=\linewidth]{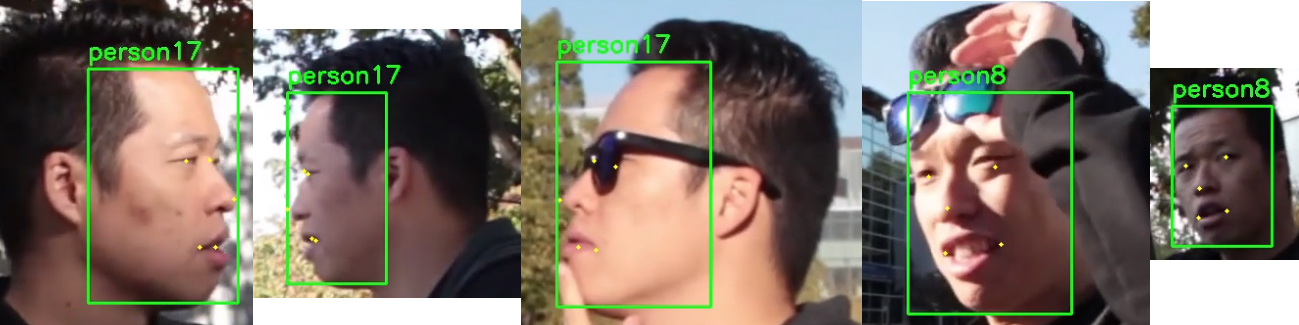}
    \caption{}
    \label{fig:fig2c}
  \end{subfigure}
  \vspace{2em}

  \begin{subfigure}{0.5\columnwidth}
    \centering
    \includegraphics[width=\linewidth]{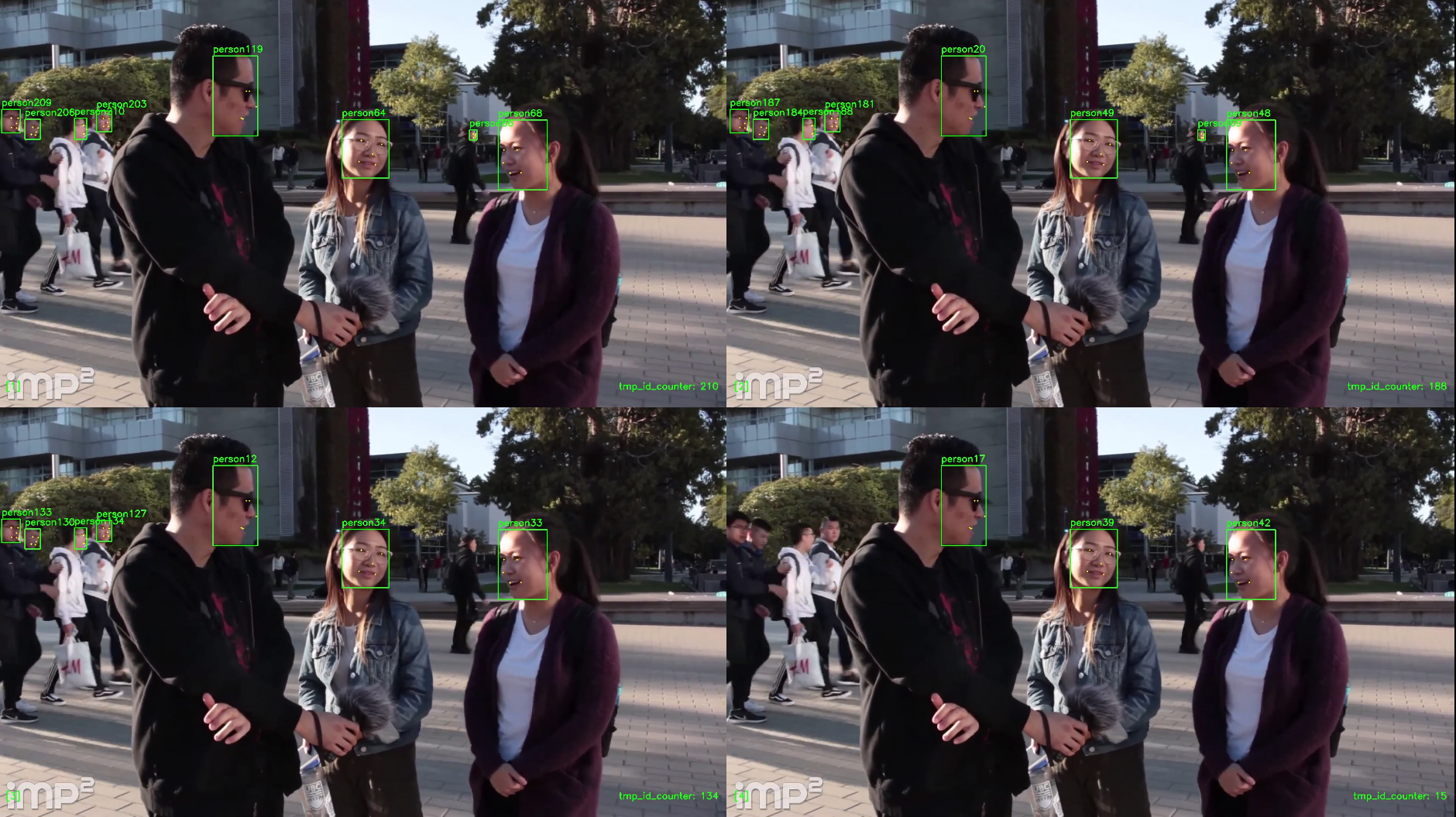}
    \caption{}
    \label{fig:fig2d}
  \end{subfigure}\hfill
  \begin{subfigure}{0.45\columnwidth}
    \centering
    \includegraphics[width=\linewidth]{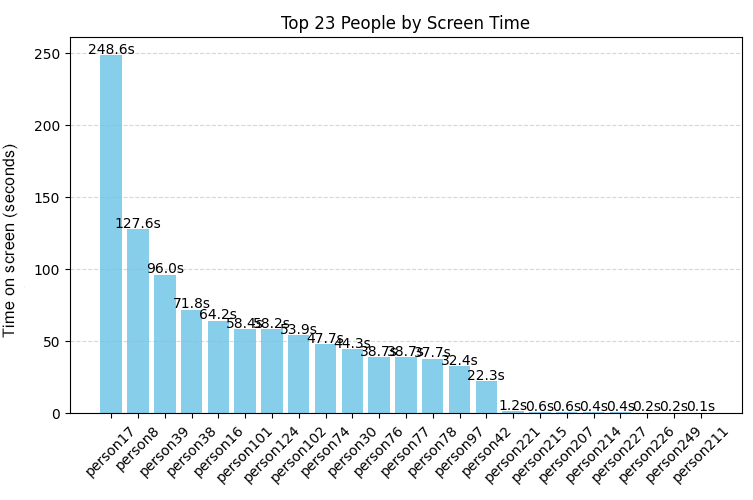}
    \caption{}
    \label{fig:fig2e}
  \end{subfigure}
    \vspace{2em}

  \captionsetup{width=0.98\linewidth}
  \caption{Video stories and face ReID analyses: (a)~face and (b)~mouth region videos;  (c)~identity mismatch,  (d)~ablation experiments on testVideo2 \cite{testVideo2}, (e)~on screen presence of all 23 identities found by the full Algorithm~\ref{alg:faceReID} in \emph{testVideo2} - the best face ReID shown in the lower-right part of image in (d).}
  \label{fig:fig2}
\end{figure}

Extensive experiments of the proposed open-set face ReID procedure were conducted based on the code implementation provided in Section~\ref{Introduction}. Experiments included visual assessment of ReID results on created visuals and video stories, statistical analysis of processing log files, and tests on three representative video sequences with varying number of people, dynamics and shooting conditions, Fig.~\ref{fig:fig2}.
 These also included side-by-side visual comparisons of the ablation experiments, Fig.~\ref{fig:fig2d}, in which the impact of each of the described algorithmic components was analyzed, Table~\ref{tab:Ablation experiments}. For more information about ablation experiments and test sequences please see the descriptions in the provided code repository. We note that the given implementation is modular and allows for independent choice of methods for each of the components in Algorithm~\ref{alg:faceReID}.

Regarding the implementation we would like to highlight the following capabilities. First, the ability to choose between near real-time generation of specified output video or efficient off-line post production of different types of video output based on detailed log of initial video analysis that is conducted only once. The log file contains information about the temporal and spatial localization of each valid identity in the gallery, their face embeddings and, depending on the settings, other data (gender and age estimates, distinctive face landmarks for each detection, and other statistics, Fig.~\ref{fig:fig2e}). Control flags and parameters allow fine-tuning of functions and their operation. Predefined video stories, with and without sound, for the user-selected identity include: a) ReID visualizations (overlays of various processing results over original video), Fig.~\ref{fig:fig2d}; b) cropped and optionally scaled face videos, Fig.~\ref{fig:fig2a}, or c) mouth region videos, Fig.~\ref{fig:fig2b}.

\vspace{1.8em}

\begin{table}[!hb]
\centering
\caption{Reduction of false identities brought by Algorithm~\ref{alg:faceReID} - ablation study}
\begin{threeparttable}
\begin{tabular}{lccccccc}
\toprule
\#identities & $\exp 1$ & $\exp2$ & $\exp3$& $\exp4$ & true & $\gamma$\tnote{a}~~[\%] &  [m:s]\tnote{b} \\
testvideo1 & 50 & 42 & 30 & 7 & 4 & 83 & 02:44\\
testvideo2 & 421 & 378 & 263 & 23 & 13 & 93 & 07:25 \\
testvideo3 & 39 & 37 & 25 & 9 & 6 & 73 & 18:45\\
\bottomrule
\end{tabular}
\begin{tablenotes}
\item[a] Alg.~\ref{alg:faceReID} relative gain:  $\gamma = (1- \exp4 /(\sum_{i=1}^{3}(\exp i)/3))\times 100\%$; \textsuperscript{b} duration
\end{tablenotes}
\end{threeparttable}
\label{tab:Ablation experiments}
\end{table}

Note that generated log files contain only anonymized data (information about face appearances), and as such are suitable for sharing even if the original videos require specific licenses.

Experimental setup was based on SCRFD face detector \cite{guo2022sample} and ArcFace embeddings \cite{deng2019arcface}, which have been trained on the subset of WebFace dataset \cite{zhu2021webface260m} and kindly provided by InsightFace project \cite{InsightFace2025}, i.e. buffalo\_l model pack\cite{buffalol2025}, with overall model size of 325 MB. VideoFace2.0 was tested on Intel i-7 notebook with 16GB of RAM and 4GB NVIDIA RTX3050 GPU. For efficient inference we have utilized ONNX runtime \cite{onnx2025} with CUDA support, instead of possibly more efficient TensorRT \cite{tensorRT2022}, and achieved a processing speed between 18-25 fps. Relatively small model size should allow for implementation on embedded platforms like Jetson Nano\cite{jetsonNano2025}, which is planned future work. For image processing and video coding we have relied on OpenCV \cite{opencv2025} and FFmpeg \cite{ffmpeg2025} libraries. In all experiments involving full Algorithm~\ref{alg:faceReID}, lower-right image in Fig.~\ref{fig:fig2d}, parameters were pre-set to $\sigma_h=0.6$, ReID similarity score threshold of $0.4$, i.e. $\tau_d=0.6$, $\tau=0.8$, and $t_{min}=60$ frames, based on extensive pre-testing. We note that these are dependant on the specific choice of components in Algorithm~\ref{alg:faceReID}.

\section{Conclusions}
\label{Conclusions}
  In this paper we have presented a novel video analysis tool for face ReID and creation of structured identity-based video outputs or video stories. The proposed algorithm proved to be robust on various challenges brought by open-set face ReID and tests corresponding to the considered  real-world application scenarios. Regarding limitations, we would like to point out that the system sometimes creates multiple identities associated with the same person, as a consequence of the proposed system design. Namely, the recognition stage in Algorithm~\ref{alg:faceReID} is relying on relatively simple identity association rule, as well as on face embeddings that do not account for occlusions and face image variations during the video. Possible solutions would be a more complex gallery updates and more robust association rules, which are planned for future work.

\section*{Acknowledgment}
{This research was supported by the Science Fund of the Republic of Serbia, project "Multimodal multilingual human-machine speech communication - \href{https://www.ktios.ftn.uns.ac.rs/ai-speak/AI-SPEAK.html}{AI SPEAK}", grant no. 7449; by the Ministry of Science, Technological Development and Innovation of the Republic of Serbia (contract no. 451-03-137/2025-03/200156); and by the \href{https://ftn.uns.ac.rs/}{Faculty of Technical Sciences}, University of Novi Sad, through the project “Scientific and artistic research work of researchers in teaching and associate positions at the Faculty of Technical Sciences,  \href{https://www.uns.ac.rs/index.php/en/}{University of Novi Sad}, 2025” (no. 01-50/295). The authors would also like to thank colleagues from Chair of Telecommunications and Signal Processing (\href{https://www.ktios.ftn.uns.ac.rs/}{KTiOS}) for fruitful discussions.}

\bibliographystyle{IEEEtranDOI}
\bibliography{VideoFace}

\end{document}